\documentclass[conference, a4paper]{IEEEtran}

\usepackage[utf8]{inputenc}
\usepackage[T1]{fontenc}
\usepackage{amsmath, amssymb, amsfonts}
\usepackage{graphicx}
\usepackage{csquotes} 
\usepackage{url}    

\usepackage{array} 

\usepackage{hyperref}
\hypersetup{
    colorlinks=true,
    linkcolor=blue,
    filecolor=magenta,
    urlcolor=cyan,
    pdftitle={Clarifying Model Transparency: Interpretability versus Explainability in Deep Learning with MNIST and IMDB Examples},
    pdfpagemode=FullScreen,
    breaklinks=true 
}

\IEEEoverridecommandlockouts

\begin{document}

\title{Clarifying Model Transparency: Interpretability versus Explainability in Deep Learning with MNIST and IMDB Examples}

\author{\IEEEauthorblockN{Mitali Raj}
\IEEEauthorblockA{\textit{School of Cyber Security and Digital Forensics} \\
\textit{National Forensic Sciences University}\\ 
New Delhi, India \\
mitali.btmtcs22215@nfsu.ac.in}
}

\IEEEpubid{XXX-X-XXXX-XXXX-X/XX/\$31.00~\copyright~2026 IEEE}

\maketitle

\IEEEpeerreviewmaketitle 
\thispagestyle{empty} 
\pagestyle{empty}    

\begin{abstract}
The impressive capabilities of deep learning models are often counterbalanced by their inherent opacity, commonly termed the "black box" problem, which impedes their widespread acceptance in high-trust domains. In response, the intersecting disciplines of interpretability and explainability, collectively falling under the Explainable AI (XAI) umbrella, have become focal points of research. Although these terms are frequently used as synonyms, they carry distinct conceptual weights. This document offers a comparative exploration of interpretability and explainability within the deep learning paradigm, carefully outlining their respective definitions, objectives, prevalent methodologies, and inherent difficulties. Through illustrative examinations of the MNIST digit classification task and IMDB sentiment analysis, we substantiate a key argument: interpretability generally pertains to a model's inherent capacity for human comprehension of its operational mechanisms (global understanding), whereas explainability is more commonly associated with post-hoc techniques designed to illuminate the basis for a model's individual predictions or behaviors (local explanations). For example, feature attribution methods can reveal \emph{why} a specific MNIST image is recognized as a '7', and word-level importance can clarify an IMDB sentiment outcome. However, these local insights do not render the complex underlying model globally transparent. A clear grasp of this differentiation, as demonstrated by these standard datasets, is vital for fostering dependable and sound artificial intelligence.
\end{abstract}


\section{Introduction} 
Deep learning (DL) has instigated transformative progress across numerous domains, such as computer vision and natural language processing, frequently demonstrating performance levels that surpass human benchmarks. Nevertheless, the sophisticated, non-linear architecture of DL models, which can encompass millions of adjustable parameters, frequently obscures their internal decision-making pathways. This "black box" attribute fuels apprehensions regarding reliability, accountability, equity, and security, particularly when deployed in critical applications.

To allay such concerns, significant research efforts have been directed towards enhancing the transparency of DL models. Central to this endeavor are two pivotal concepts: \textbf{interpretability} and \textbf{explainability}. Despite their common interchangeable usage, a more detailed examination uncovers crucial distinctions. This paper endeavors to analyze and contrast these notions, anchoring the discussion with tangible examples derived from established DL applications: image recognition on the MNIST dataset and text-based sentiment classification using the IMDB dataset. These case studies will effectively illustrate how specific prediction explanations can be achieved even when a model's overall global interpretability remains limited.

\section{Defining Interpretability and Explainability}
No universally endorsed definitions exist for interpretability or explainability, which has led to a degree of terminological fluidity in academic discourse. Nonetheless, practical definitions can be formulated based on prevalent usage and influential research.

\subsection{Interpretability}
\begin{itemize}
\item \textbf{Definition:} The notion of interpretability centers on the extent to which a human observer can grasp the \textit{causal relationships} or the \textit{underlying functional principles} governing a model's decision-making (Lipton, 2018 \cite{Lipton2018}; Doshi-Velez \& Kim, 2017 \cite{DoshiVelezKim2017}). The emphasis is on comprehending \textit{how} the model functions at a systemic level.
\item \textbf{Focus:} Intrinsic characteristics of the model, architectural transparency, global understanding.
\item \textbf{Goal:} The construction of models that are fundamentally comprehensible to humans, whether in their complete form (global interpretability) or regarding their individual components.
\item \textbf{Example Question:} \enquote{By what general process does this Convolutional Neural Network (CNN) convert raw pixel data, through its successive layers, into a final digit classification?}
\end{itemize}

\subsection{Explainability (within XAI)}
\begin{itemize}
\item \textbf{Definition:} Explainability pertains to the capacity to furnish an interface or supplementary data that elucidates a model's operational behavior, especially its individual predictions, in a manner that is intelligible to humans (Arrieta et al., 2020 \cite{Arrieta2020}). This often entails the application of post-hoc methods to pre-trained models.
\item \textbf{Focus:} Justification of model outputs after the fact, understanding at the level of individual predictions, local explanations.
\item \textbf{Goal:} The generation of clear rationales for specific instances or for the model's general behavioral patterns, frequently without needing to make the model's entire internal architecture transparent.
\item \textbf{Example Question:} \enquote{What factors led the model to classify \textit{this particular MNIST image} as a '3' rather than an '8'?} or \enquote{Which specific words in \textit{this IMDB review instance} were most decisive for its 'positive' sentiment label?}
\end{itemize}

\section{Comparative Analysis}
The principal distinctions between interpretability and explainability are outlined in Table \ref{tab:comparison}.

\begin{table}[htbp]
\caption{Distinguishing Features of Interpretability and Explainability (XAI)}
\label{tab:comparison}
\centering
\begin{tabular}{|p{0.2\columnwidth}|p{0.35\columnwidth}|p{0.35\columnwidth}|}
\hline
\textbf{Attribute}      & \textbf{Interpretability}                                 & \textbf{Explainability (XAI)}                                  \\
\hline
\textbf{Core Objective} & Comprehend the model's internal logic/system (Global Understanding). & Understand the rationale for a specific decision/prediction (Local Explanation). \\
\hline
\textbf{Methodological Timing} & Primarily intrinsic to model design (ante-hoc).      & Frequently extrinsic, applied post-training (post-hoc).        \\
\hline
\textbf{Preferred Model Type} & Leans towards simpler, inherently transparent architectures. & Applicable to complex, opaque models (typical of DL).  \\
\hline
\textbf{Reach of Understanding} & Can be global (entire model) or modular (components). & Often instance-specific (local predictions), though global approximations are possible. \\
\hline
\textbf{Role of Human User}   & Human can directly understand the model's functioning. & Human receives an explanation produced by an auxiliary process.   \\
\hline
\textbf{Exemplary Techniques} & Simpler models (e.g., linear regression, decision trees), rule induction, concept-based systems (e.g., ProtoPNet). & Feature attribution maps (e.g., Grad-CAM), LIME, SHAP, counterfactual reasoning, attention visualization. \\
\hline
\textbf{Interrelation}& An interpretable model is by nature explainable.      & An explainable model is not necessarily interpretable.       \\
\hline
\end{tabular}
\end{table}

\section{Illustrative Case Studies: Anchoring Explainability Concepts}
To provide a concrete illustration of this distinction—especially the point that explainability (yielding local explanations) can be attained for opaque models lacking comprehensive interpretability (global understanding)—we investigate two prevalent deep learning scenarios.

\subsection{Case Study 1: MNIST Handwritten Digit Recognition}

\begin{figure}[t] 
\centering 
\includegraphics[width=\columnwidth]{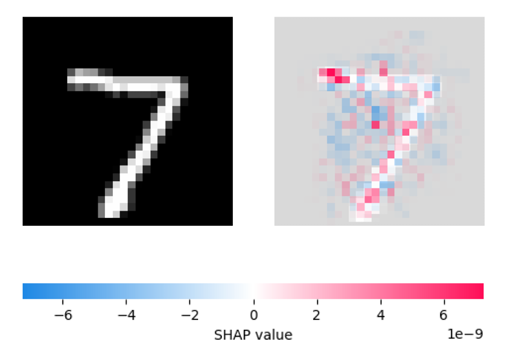}
\caption{SHAP-based local explanation of a CNN's prediction for an MNIST digit classified as '7'. Left: Original input image. Right: SHAP heatmap indicating pixel-level feature contributions; red areas contribute positively to the prediction '7', while blue areas contribute negatively.}
\label{fig:mnist}
\end{figure}

The MNIST dataset consists of 70,000 grayscale images depicting handwritten digits (0-9). CNNs represent the standard, high-efficacy models for this recognition task.

\begin{itemize}
    \item \textbf{The Opaque Nature of the Model:} A standard CNN for MNIST typically incorporates several convolutional and pooling layers, followed by fully-connected layers, amounting to potentially millions of parameters. Although the architectural blueprint is human-designed, deciphering precisely how a particular configuration of learned filter weights across all layers culminates in a digit classification presents a formidable challenge. The model's global operational mechanics remain largely obscure, precluding an easy articulation of its learned comprehensive rule-set for achieving global understanding.
    
    \item \textbf{Attaining Local Explainability (Post-hoc):} Notwithstanding the CNN's opacity, explanations for individual predictions are obtainable.
    \begin{itemize}
        \item \textbf{Feature Attribution Maps (e.g., Grad-CAM \cite{Selvaraju2017}, SHAP):} For an image classified as a '7', such methods can produce a visual heatmap or assign scores that pinpoint the input image pixels most critical to this '7' determination (e.g., the horizontal bar and diagonal stroke characteristic of a '7'). This offers a localized, post-hoc rationale: \enquote{The model designated this as '7' due to its focus on these '7'-indicative pixel areas.} (See Fig.~\ref{fig:mnist}).
        \item \textbf{LIME \cite{Ribeiro2016}:} LIME can locally approximate the CNN's decision boundary for a given image by introducing perturbations (like toggling superpixels) and training a simpler, understandable model (such as a sparse linear regressor) on these variations. This would identify which superpixels steered the decision towards '7'.
    \end{itemize}
    
    \item \textbf{Takeaway from MNIST Analysis:} XAI tools empower us to effectively generate a local \textit{explanation} for why a specific image is assigned a particular digit label by highlighting crucial input features. Nevertheless, these explanations do not transform the entire CNN into a globally \textit{interpretable} system. We gain insight into the 'why' for an individual instance, but not necessarily the complete 'how' of the model's overarching learned function.
\end{itemize}

\subsection{Case Study 2: IMDB Movie Review Sentiment Analysis}

\begin{figure}[t]
\centering
\includegraphics[width=\columnwidth]{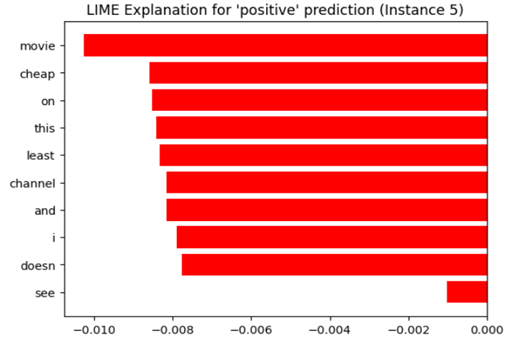}
\caption{LIME explanation for a positive sentiment prediction on an IMDB movie review. Although the predicted label is positive, the words shown (e.g., "cheap," "least," "channel") had negative contributions to the classification. The explanation helps interpret the model's local reasoning for this instance.}
\label{fig:imdb}
\end{figure}

The IMDB dataset comprises 50,000 movie reviews, each categorized as positive or negative. Models like RNNs (e.g., LSTMs) and Transformers (e.g., BERT) deliver leading performance on this task.

\begin{itemize}
    \item \textbf{The Opaque Nature of the Model:} Architectures such as LSTMs or Transformers employ intricate sequential data processing and attention mechanisms on input word embeddings. Tracing the transformation of an input sentence's semantic content through recurrent states or multiple self-attention layers to a final sentiment score is a complex endeavor. The learned representations are high-dimensional, and their precise semantic interpretations are not always readily apparent, making these models inherently uninterpretable for global understanding.
    
    \item \textbf{Attaining Local Explainability (Post-hoc):}
    \begin{itemize}
        \item \textbf{Attention Visualization (Relevant for Transformers):} When utilizing a Transformer, attention weights can reveal which words the model emphasized while processing others, or which words received the most attention from the final classification token. For a review classified as 'positive', strong attention might fall on terms like \enquote{excellent,} \enquote{brilliant,} or \enquote{loved.}
        \item \textbf{LIME \cite{Ribeiro2016} or SHAP \cite{LundbergLee2017}:} These techniques can be adapted for text models, for example, by omitting words from a review and observing the consequent shift in prediction probability. They can then allocate importance scores to each word in a specific review, signifying which words were most instrumental in the 'positive' or 'negative' classification. An example explanation might be: \enquote{This review was classified as negative primarily due to the substantial negative influence of the words 'awful' and 'waste'.}
    \end{itemize}
    
    \item \textbf{Takeaway from IMDB Analysis:} Analogous to the MNIST scenario, XAI methods enable the generation of local \textit{explanations} for the sentiment prediction of an individual review by identifying key influential textual elements. This yields valuable, localized comprehension. However, the comprehensive internal logic by which the LSTM or Transformer model processes language to discern sentiment remains multifaceted and largely uninterpretable from a global perspective.
\end{itemize}

\section{Synthesizing Insights from Case Studies}
The analyses of MNIST and IMDB datasets furnish concrete evidence that:
\begin{itemize}
    \item Deep learning models, despite their power, frequently function as opaque systems, rendering their global internal processes challenging to interpret for holistic understanding.
    \item Post-hoc XAI methodologies can deliver valuable \textit{local explanations} for specific predictions generated by these opaque models. Pixel attribution maps for MNIST images and word-level importance scores for IMDB reviews serve as excellent examples.
    \item These local explanations address the \enquote{why this particular prediction?} query, thereby enhancing localized trust and comprehension.
    \item Nonetheless, the provision of such local explanations does not inherently render the entire model globally \textit{interpretable}. The sophisticated underlying mechanisms often remain inscrutable.
    \item Consequently, a model that is explainable (i.e., for which local justifications for outputs can be produced) is not axiomatically interpretable (i.e., one whose fundamental operations are transparent for global understanding).
\end{itemize}

\section{General Techniques in Deep Learning}
The field offers a spectrum of methods, extending beyond those highlighted in our case studies.

\subsection{Interpretability-focused Techniques (Often "Interpretable by Design" for Global Understanding)}
\begin{itemize}
\item \textbf{Structured Attention Mechanisms:} Designing models where attention is a constrained, core element can foster interpretability by forcing explicit focus indication.
\item \textbf{Concept-based Architectures (e.g.,\ TCAV):} These aim to link model decisions to high-level, human-intelligible concepts.
\item \textbf{Generalized Additive Models (GAMs) / Neural Additive Models (NAMs):} These learn a sum of non-linear transformations of individual input features.
\item \textbf{ProtoPNet (Prototypical Part Network):} Predictions are made by comparing input segments to learned class-specific prototypes.
\item \textbf{Rule Extraction from Neural Networks:} This involves attempts to translate the learned knowledge of a neural network into a set of symbolic rules.
\end{itemize}

\subsection{Explainability-focused Techniques (Often Post-hoc for Local Explanations)}
\begin{itemize}
\item \textbf{Saliency/Feature Attribution Maps (as in MNIST study):}
\begin{itemize}
\item \textbf{Gradient-derived:} Methods like Vanilla Gradients, Integrated Gradients, Grad-CAM.
\item \textbf{Perturbation-based:} Techniques such as occlusion sensitivity, or certain SHAP algorithm variants.
\end{itemize}
\item \textbf{LIME (Local Interpretable Model-agnostic Explanations) (relevant to MNIST \& IMDB studies):} This approach locally trains a simpler, interpretable surrogate model.
\item \textbf{SHAP (SHapley Additive exPlanations) (relevant to IMDB \& MNIST studies):} A game-theoretic method for assessing feature importance.
\item \textbf{Counterfactual Explanations:} These identify the minimal input modifications required to change a prediction.
\item \textbf{Influence Functions:} These assess the impact of individual training samples on model parameters and predictions.
\end{itemize}

\section{Importance and Applications}
Both interpretability (which targets global understanding) and explainability (which provides local explanations) are vital for endeavors such as building trust, facilitating model debugging, ensuring fairness, meeting regulatory compliance, enabling scientific discovery, and guaranteeing safety in critical systems. The case studies underscore that even when complete global interpretability is unattainable, instance-specific local explainability offers significant value.

\section{Challenges and Limitations}
\begin{itemize}
\item \textbf{Performance vs. Transparency Trade-off:} High-accuracy DL models (as used for MNIST/IMDB) frequently lack global interpretability. Simpler, more interpretable alternatives might compromise predictive power. Post-hoc local explanations are often approximations.
\item \textbf{Veracity of Explanations:} Validating that local explanations genuinely mirror the model's reasoning for a specific instance remains a significant hurdle.
\item \textbf{Human-Centricity of Explanations:} The definition of an effective or \enquote{understandable} local explanation is inherently subjective and varies with context and user.
\item \textbf{Objective Evaluation:} Quantifying the quality of global interpretability or a local explanation poses considerable difficulty.
\item \textbf{Resource Intensiveness:} Many XAI techniques demand substantial computational resources.
\item \textbf{Breadth of Insight:} Most XAI tools deliver local explanations. Achieving a reliable global understanding of entire DL models is a more elusive goal.
\end{itemize}

\section{The Symbiotic Relationship}
Although distinct, interpretability (global understanding) and explainability (local explanations) are not mutually exclusive concepts.
\begin{itemize}
\item Incorporating inherently interpretable modules can enhance the coherence and reliability of post-hoc local explanations.
\item Local explanation methods can serve as auditing tools for models that are partially interpretable.
\end{itemize}
The overarching aim is to achieve an adequate level of understanding pertinent to the specific application. This might be realized through an intrinsically globally interpretable model or through an opaque model supplemented by robust local explanations, as demonstrated by our MNIST/IMDB scenarios where local explanations furnish key insights despite the models' overall complexity.

\section{Future Directions}
\begin{itemize}
\item \textbf{Development of Inherently Interpretable Deep Learning Systems:} Pursuing novel DL architectures that combine high performance with inherent transparency for global understanding.
\item \textbf{Advancement of Causal Explanations:} Progressing from correlational feature importance in local explanations towards insights grounded in causal inference.
\item \textbf{Creation of Interactive and Personalized Explanation Interfaces:} Designing tools that provide tailored local explanations responsive to user queries.
\item \textbf{Enhancing Robustness and Scrutiny of Explanations:} Ensuring that local explanations are resilient to manipulation and are not deceptive.
\item \textbf{Establishment of Standardization and Benchmarking Protocols:} Developing common evaluative frameworks for both types of methodologies.
\end{itemize}

\section{Conclusion}
Within the deep learning landscape, \textbf{interpretability} is primarily associated with the inherent comprehensibility of a model's internal operations, striving for a \emph{global understanding} of its systemic behavior. This is commonly pursued via architectural choices that promote transparency. Conversely, \textbf{explainability} (XAI) generally entails the use of post-hoc methodologies to offer human-intelligible rationales for specific model outputs, thereby delivering \emph{local explanations} concerning the behavior of intricate, often opaque, systems.

Our investigations using the MNIST and IMDB datasets, coupled with the practical application of XAI utilities, serve to concretely delineate this difference. For a Convolutional Neural Network (CNN) tasked with MNIST digit classification, explainability techniques such as SHAP or other saliency approaches can produce \textbf{pixel importance maps}. These visualizations identify the pixels most critical to a particular digit's classification (e.g., elucidating \emph{why} a given image is identified as a '7' by highlighting its formative pixel configurations). In a similar vein, for models like LSTMs or Transformers engaged in IMDB movie review analysis, techniques including LIME or attention mechanisms can yield \textbf{word importance scores}. These scores clarify a sentiment assessment by pinpointing the most influential words or phrases (e.g., explaining \emph{why} a review is classified as 'positive' through specific textual cues).

In both instances, while these post-hoc local explanations furnish valuable understanding of \emph{why} an individual prediction was rendered, the overarching, sophisticated decision-making architecture of these high-performing deep learning models largely remains uninterpretable. The local explanations, by concentrating on input feature attributions for specific outcomes, supply justifications but do not provide a holistic, global grasp of the model's complete learned logic or its internal operational transformations.

Consequently, although a globally interpretable model is inherently explainable at both local and global scales, a model that is merely explainable—as illustrated by standard DL applications to MNIST and IMDB where local explanations for distinct predictions can be generated—is not necessarily globally interpretable. This distinction holds significant importance. Depending on post-hoc local explainability for complex models is frequently a pragmatic approach; however, recognizing its constraints concerning genuine global model comprehension is essential. Progress in artificial intelligence demands advancements on both fronts: the creation of more intrinsically interpretable models for achieving global understanding, and the refinement of robust explainability techniques for trustworthy local justifications. Such dual development is paramount for cultivating accountable, beneficial, and reliable AI systems.

\end{document}